# Solar potential analysis over Indian cities using high-resolution satellite imagery and DEM.

**Jai G. Singla[1]\*, Atharva Kalase[2]**

*Space Applications Centre[1], ISRO, Ahmedabad – 380015, India*

*Vishwakarma institute of Techonology[2], Pune, India*

E-mail: jaisingla@gmail.com \*Corresponding Author

**Abstract:** Most of the research work in the solar potential analysis is performed utilizing aerial imagery, LiDAR data, and satellite imagery. However, in the existing studies using satellite data, parameters such as trees/ vegetation shadow, adjacent higher architectural structures, and eccentric roof structures in urban areas were not considered, and relatively coarser-resolution datasets were used for analysis. In this work, we have implemented a novel approach to estimate rooftop solar potential using inputs of high-resolution satellite imagery (0.5 cm), a digital elevation model (1m), along with ground station radiation data. Solar radiation analysis is performed using the diffusion proportion and transmissivity ratio derived from the ground station data hosted by IMD. It was observed that due to seasonal variations, environmental effects and technical reasons such as solar panel structure etc., there can be a significant loss of electricity generation up to 50%. Based on the results, it is also understood that using 1m DEM and 50cm satellite imagery, more authentic results are produced over the urban areas.

## 1   Introduction

Since ancient times, the Sun is considered a sacred God and has been worshipped as a life-giver. According to the ministry of New and Renewable Energy the Government of India, India is endowed with vast solar potential which is around 5,000 trillion kWh per year (MNRE website). The National institute of solar energy has assessed the country's solar potential of about 748 GW assuming 3% of the wasteland area to be covered by Solar PV modules (MNRE website). Indian Government has also emerged with some government subsidy schemes to empower the solar potential of India such as "Kisan Urja Suraksha evam Utthan Mahaabhiyaan (Kusum Scheme)", "International Solar Alliance", "One Sun, One World, One Grid (OSOWOG)", "National Solar Mission", "Rooftop Solar Scheme" and "National Wind Solar Hybrid Policy 2018". In order to properly assess solar potential, there is a requirement to calculate solar potential of a region before the installation of solar panels. Remote sensing satellite data serves as a very good solution to calculate the solar potential over a larger geographical area. Due to availability of high resolution satellites and automatic weather stations, solar potential can be assessed quite accurately over dense urban areas too.
 Many researchers explored the use of remote sensing data in solar energy calculation in the past. J. A. Ruiz-Arias et al (2009) performed a comparative analysis on high-profile hilly topographic areas using four different solar potential analyst's toolsets such as Solar Analyst, SRAD, r.Sun, and Solei -32 using 20m and 100m DEM. Their results showed that solar radiation variability could be reasonably estimated under clear sky conditions using DEM models with 100m of spatial resolution. R. Singh et al (2013) proposed a micro-synthesized stimulation methodology to analyze solar potential using PVsyst software over the IIT Bombay campus. H. Sogukpinar et al. (2016) made use of PVGIS (PVG website) software and performed a comparative analysis of software output data with experimental data acquired by Govt. authorities in Turkey. It was analyzed that the region of South-eastern Anatolia has the maximum solar potential followed by the Mediterranean coast. V. K. Mahaver et al (2018) performed theoretical analysis using PVsyst, SAM, and RET software simulations to predict power generation in Rajasthan. V. M. V. Nanda et al (2020) analyzed the shade analysis under trees according to seasons and suggested better parking places using LiDAR and DEM data of resolution 0.5m with r.Sun and GRASS GIS. D. Kumar et al (2020) worked on solar resource variability analysis at a spatial scale in the southern states of India. This work utilized satellite-derived model solar insulation datasets. Kausika, B.B et al (2021) worked on calibrating and validating ArcGIS Pro's solar radiation toolset for assessing the photovoltaic potential in the Netherlands. They assessed data from various weather stations in the city and compared it with results using a solar radiation toolset, which was calibrated and validated accordingly. S. Kolecansky et al (2021) worked on 3D and 2D solar potential analysis using 2D (r.Sun) and 3D (v.Sun) solar radiation models over a city area. The 3D v.Sun solar radiation model predicted solar irradiances on vertical surfaces with much better accuracy compared to 2D models. A. O. Salau et al (2021) proposed an Angstrom Prescott model on a horizontal surface using Sunshine statistics from NASA over Ethiopia. T. S. Anandh et al (2022) performed future wind and solar potential analysis for the past 55 years where simulations consisted of CORDEX-SA, 13 CMIPS, and 13 CMIP6 models, which displayed loss in solar potential in the region of the Southern and north-western part of India. All the above-mentioned studies either considered only the 2D satellite data for solar radiation assessment or worked with coarser resolution DEM (30-100m). It has also been observed that all the past studies are based mainly on software simulations such as PVsyst, r.Sun, v.Sun, and PVGIS. Most of these SWs perform solar radiation calculations without taking the height parameter (DEM) into account. v.Sun is only the 3D variant of r.Sun that also takes care of new vector-voxel calculation procedures for complex 3D urban surfaces. v.Sun uses a linked turbidity factor and beam diffuse radiation coefficients, which require readings from nearby weather stations. Whereas the method used in our work uses simplified models and provides a very detailed view of shading and slopes using very high-resolution data.

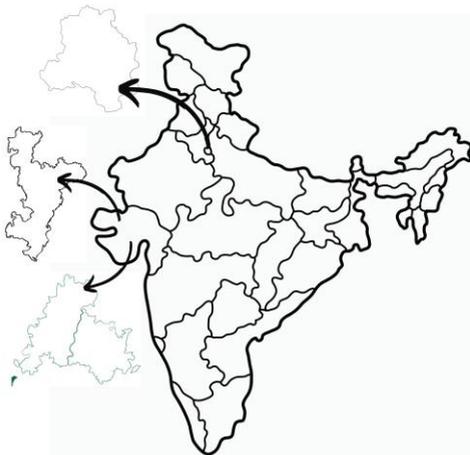

Fig. 1. Geographical Site Analysis Map

Considering all the literature surveys, there is good scope to study solar potential on building rooftops in urban regions using very high-resolution satellite 3D data over the Indian region. The main objectives of our study are:

1. Calculation of solar radiation values over each rooftop at building level over the city area using very high-resolution DEM and imagery.
2. Consideration of all important parameters such as diffusion and transmittivity, solar radiation direction, slope, and nearby structures for calculations of solar radiations.

To accomplish the above-mentioned objectives, the following contribution is made in this study:

i. A new approach is devised to calculate solar radiation value over urban/city areas by making use of very high-resolution satellite imagery(50cm) and the Digital Surface Model(1m), where the height parameter of the surface and rooftops have been considered for solar potential analysis.

ii. The parametrical analysis is done on the data where the rooftop slope, the direction of the

iii. Consideration of sky transmissivity and diffusivity parameters for the calculation of direct and diffuse insolation. Diffusion ratio and transmissivity values were derived based on IMD data for different sites for the assessment and validation of solar radiation values.
  iv. In addition to this, we also carried out the comparisons among 2D solar radiation analysis using only raster 2D data and 3D data with resolutions of 1m, 10m, and 30m respectively.
  v. This study has several unique research elements and thus it is one of the unique studies of its kind in India as per our knowledge.

## 2 Geographical Site Analysis

Ahmedabad is one of the most well-known and former capitals of Gujarat. Here, we have a variety of buildings where the height of buildings ranges from low-ground bungalows to very high skyscrapers, also the average temperature in Ahmedabad ranges from $12^0$ C to $42^0$ C in the first half and $13^0$ C to $36^0$ C in the second half of the year. Here the average daylight time is 11 to 13.5 hours, also the geographical location is closer to the equator where the Sunlight is incident perpendicular compared to other places which make it an ideal site for analysis. Gandhinagar is located North-Eastern-Central part of Gujarat which is also known as the "Tree Capital of India" because of the greenery all around. The average temperature in Gandhinagar ranges from $14^0$ C to $42^0$ C in the first half and $13^0$ C to $36^0$ C in the second half of the year and the average daylight time is 11 to 13.5 hours. Delhi the capital of India is spread over a vast area of 1486 square kilometers experiencing diverse weather states throughout the year. The average temperature in Delhi ranges between a minimum of $9.2^0$ C to $32^0$ C and a maximum of $20^0$ C to $43^0$ C where a gradual increase is observed from January till June and in the second half a drastic decline is observed.

## 3 Study Area & Dataset used.

In this research study, the Digital Elevation Model (DEM) of varied spatial resolution is used i.e., DEM with 1m, 10m, and 30m grid intervals as per the details mentioned in Table 1. All the Digital Elevation Models are extracted using the approach of Satellite Photogrammetry from raw stereo pairs captured using Remote Sensing (RS) satellites. Most of the high-resolution 1m DEM datasets are procured from AW3D whereas 10m DEM is acquired through the internal DP team at SAC. The area over Ahmedabad, Gandhinagar, and Delhi are taken as Areas of Interest (AOI). Ahmedabad, Gandhinagar, and Delhi are one of the most populated cities in India with an area of 505, 326, and 1486 sq. km respectively. DEMs over AOI contain sufficient urban areas and numerous numbers of low/ medium and high-rise buildings with different terrain complexity to conduct our research work.

| Sr No | Area | Width, height | Pixel size | Sensor Information and Data Type |
|---|---|---|---|---|
| 1. | Ahmedabad | 12000, 12000 | 1 m | DEM generated using stereo images from World view -2, RMSE 2m |
| 2. | Ahmedabad | 6880, 8824 | 10m | Cartosat-1 DEM, RMSE 7m |
| 3. | Ahmedabad | 3601, 3601 | 30m | Aster 30M free DEM, RMSE ~8m |
| 4. | Gandhinagar | 7088,8446 | 1m | AW3D 1m DEM, RMSE 3m |
| 5. | Gandhinagar | 6880, 8824 | 10m | Cartosat-1 DEM, RMSE 7m |
| 6. | Gandhinagar | 3601, 3601 | 30m | Aster 30M DEM, RMSE 8m |
| 7. | Delhi | 4977, 6492 | 1m | AW3D 1m DEM, RMSE 3m |
| 8. | Delhi | 6880, 8824 | 10m | Cartosat-1 DEM, RMSE 7m |
| 9. | Delhi | 3601, 3601 | 30m | Aster 30M DEM, RMSE 8m |
| 10. | Ahmedabad, Gandhinagar, and Delhi | 12000, 12000 7088, 8446 4977, 6492 | 50cm | Worldview 2 image data |

*Table -1: Detailed data description*

**Global radiation and diffused radiation datasets:**

The India Meteorological Department (IMD) has hosted the diffused and global radiation data available through ground weather stations over Ahmedabad, Gandhinagar, and Delhi. This data is helpful in various commercial as well as research analyses. Solar diffusion is referred to as the radiation being scattered away by the earth's atmosphere. These values are affected by various parameters such as cloud cover, pollution, fog, and aerosol density. Global radiation denotes the total radiation value entering the earth's environment (both direct and diffused), The IMD data for worldwide radiation is critical for analyzing the availability of solar energy resources and estimating the potential for solar power generation.

The IMD statistics on solar diffusion and global radiation values for Ahmedabad, Gandhinagar, and Delhi provide a complete insight into the solar radiation patterns peculiar to these sites. This information is typically provided at hourly, daily, monthly, or yearly intervals and may contain factors such as solar radiation intensity, cloud cover, Sunshine duration, and atmospheric conditions. The monthly Average of global radiation data and diffuse radiation are listed in Table 2.

| Time series | City | No. of hours used for analysis | Average Global radiation data | Average Diffuse radiation data |
|---|---|---|---|---|
| Jan-21 | Ahmedabad | 16 | 0.385 | 0.779 |
| Feb-21 | Ahmedabad | | 0.365 | 1.108 |
| Mar-21 | Ahmedabad | | 0.476 | 1.338 |
| Apr-21 | Ahmedabad | | 0.596 | 1.456 |
| May-21 | Ahmedabad | | 0.500 | 1.000 |
| Jun-21 | Ahmedabad | | 0.759 | 1.256 |
| Jul-21 | Gandhinagar | | 0.667 | 0.977 |
| Aug-21 | Gandhinagar | | 0.680 | 1.000 |
| Sep-21 | Gandhinagar | | 0.599 | 0.728 |
| Jan-21 | Delhi | | 0.498 | 0.634 |
| Feb-21 | Delhi | | 0.857 | 0.915 |
| Mar-21 | Delhi | | 0.783 | 1.111 |

*Table -2: Monthly average of global radiation and diffuse radiation (source: IMD)*

## 4 Methodology

The methodology delineated in Figure 2 involves a systematic approach aimed at effectively harnessing solar energy potential from rooftops. The methodology is divided in to four major sections: data pre-processing, generation of slope and aspect layer, generation of solar radiation raster, criteria for filtering of rooftops and calculation of power per building. It begins with a pivotal phase of data preprocessing, wherein a Digital Elevation Model (DEM) is utilized to derive building footprint shapes for targeted areas, specifically Ahmedabad, Gandhinagar, and Delhi. These footprints provide a foundational outline for rooftop surfaces, crucial for subsequent solar radiation calculations.

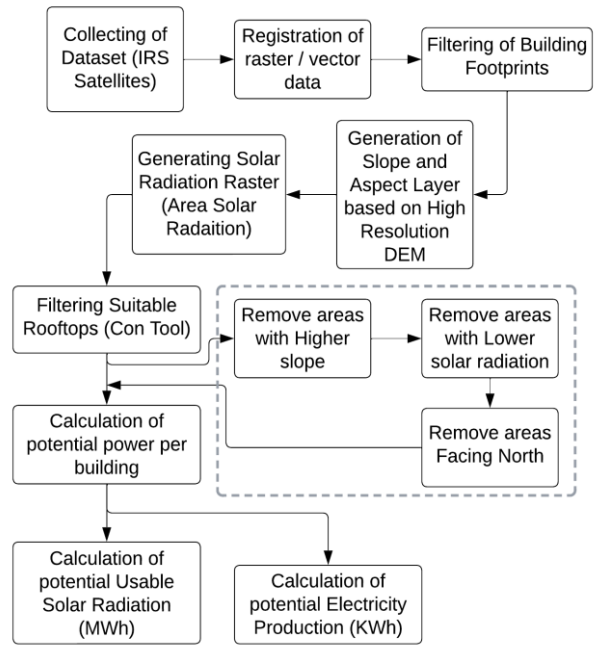

*Fig. 2. Flow Chart of Methodology*

Following data preprocessing, the methodology proceeds to generate slope and aspect layers through a surface parameter methodology employing a spatial analyst model. These layers furnish insights into the inclination and orientation of rooftops, facilitating the identification and filtration of viable candidates for solar panel installation. Subsequently, a solar radiation layer is developed using the Area Solar Radiation toolset, incorporating customized parameters derived from pertinent data, such as insolation and transmissivity data sourced from the India Meteorological Department (IMD).

However, not all rooftops are equally suitable for solar panel deployment, owing to variations in orientation and shading. Thus, a set of criteria is established to discern rooftops primed for optimal electricity generation. Roofs deemed unsuitable are systematically excluded from further consideration. Moreover, leveraging a self-defined formula tailored for electricity generation calculations, the methodology quantifies the potential energy output for each selected building, thereby enabling informed decision-making regarding solar energy integration. In essence, this methodological framework represents a comprehensive and iterative process designed to maximize solar energy utilization from rooftops, amalgamating spatial analysis techniques with customized criteria to identify and prioritize viable candidates for solar panel deployment.

### 4.1. Data preprocessing

For analysis, the Digital Elevation Model Raster (DEM) file (.tiff) and Building footprint shapefile (.shp) are used as input. A Digital Elevation Model represents the elevation of location through a cartographic dataset where each cell visualizes the elevation in the form of shades of colors. Each cell has parameters in the form of 'X & Y' which represents the surface location of a point on land and 'Z' which represents the elevation profile value of a distinct location in a distinct pixel. The spatial resolution is an important factor where in a coarser resolution of 1m DEM the elevation of a 1m*1m square on land is represented by one cell. Cell is symbolized as darker gray cells represent low height values and brighter ones represent more height values. Building Footprint files are generated automatically using a deep learning model (Unet+ VGG19) with an accuracy of 92% (Singla et al, 2023) using the raster datasets. Building footprints along with DEM data depict the outline of the building on the ground with parameters such as the height of the building in meters, category, area of the rooftop, and no of floors.

### 4.2. Generation of slope and aspect layer.

To understand the nature of the rooftop surface, the slope layer, and aspect layers are generated using a spatial analyst model tool. The aspect layer defines the direction of the slope of the surface, the direction is defined in degrees, which is depicted by hue color. Slope defines the rate of change of slope according to DEM values. Both play an important role in filtering the higher slope regions and north-facing slope regions.

When conducting slope calculation on a digital elevation model (DEM), the resulting output signifies the rate of change of elevation for each cell within the DEM. Similarly, when aspect calculation is performed on a DEM, it provides information regarding the compass direction that the downhill slope faces at each specific location. This information is expressed in positive degrees ranging from 0 to 360, measured clockwise from the north direction. The surface parameter computes slope and aspect concerning neighboring cells, it calculates using cells of dimension ranging from 3 x 3 to 15 x 15. The geodesic method was used to calculate the slope values as it gives accurate results on the ellipsoidal geometry surface.

### 4.3. Generation of solar radiation raster

The Area Solar Radiation approach uses a Digital Elevation Model (DEM) as its major input to compute solar radiation throughout a particular terrain. This procedure makes use of a sophisticated model that considers a variety of elements that influence solar irradiance, such as the Sun's position at different times of the year and day, potential obstacles such as surrounding structures or plants, and the surface's slope and orientation. The high-resolution DEM contains critical information about the landscape's properties, such as the existence of obstacles, surface orientation, and slope angles. These factors are critical in effectively predicting the interaction of Sunlight and topography. Area solar radiation process generates a raster layer, with each cell representing a distinct place on the terrain. The cell values represent the quantity of solar radiation received at that position, measured in watt-hours per square meter (Wh/m2). Detailed methodology to generate solar radiation raster is explained under sub-sections.

#### 4.3.1. Solar Radiation Equation Calculation

The sum of direct solar radiation and diffuse solar radiation nomenclature as "Dirtot" & "Diftot" respectively results in Global Solar radiation "Globaltot" considering Sun map and sky map sectors.

**Direct radiation (Dirtot)** is calculated for each Sun map sector using a transmission model that considers gap fraction, Sun position, atmospheric attenuation, and ground receiving surface orientation. The total direct radiation received at this location is calculated by adding the radiation from all sectors. **Diffuse insolation (Diftot)** for a given location is calculated using the viewshed raster and a sky map of the study area. The sky map depicts a hemispherical view of the entire sky, divided into sky sectors defined by zenith and azimuth angles. Each sector is assigned a unique identifier, as well as the centroid, zenith, and azimuth angles.

$$Global_{tot} = Dir_{tot} + Dif_{tot} \quad \ldots Eq\ 1$$

Where,
$$Dir_{tot} = \Sigma\, Dir_{\theta,\alpha} \quad \ldots Eq\ 2$$

To calculate the direct Sun insolation having a direct Sun map sector (Dirθ, α) with parameter "θ" as the zenith angle and α as the azimuth angle the following formula is applied:

$$Dir_{\theta,\alpha} = SConst * \beta^{m(\theta)} * SunDur_{\theta,\alpha} * SunGap_{\theta,\alpha} * \cos(AngIn_{\theta,\alpha}) \quad \ldots Eq\ 3$$

Where:
- β – represents the transmissivity factor of the atmosphere for the shortest path.
- SConst is the solar flux outside the atmosphere at the mean Earth-Sun distance, also known as the solar constant. The solar constant utilized in this analysis is 1367 W/m2. This is compatible with the World Radiation Center's (WRC) solar constant.
- The relative optical path length, m(θ), is measured as a proportion to the zenith path length.
- SunDurθ,α — Time duration of the sky sector. For most sectors, it is calculated by multiplying the day interval (e.g., month) by the hour interval (e.g., half an hour). The duration of partial sectors near the horizon is computed using spherical geometry.
- SunGapθ,α: The gap fraction for the Sun map sector.
- AngInθ,α is the angle of incidence between the centroid of the sky sector and the axis normal to the surface.

### 4.3.2. Selection of importation parameters for Solar radiation

Following important parameter were selected for generation of solar radiation layer.
- Input Raster – DSM (Input is given as 1m/10m or 30m DEM) / and 2D image.
- Time configuration – Whole year **(Annual)** / Particular Day (Monthly).
- Hour Interval – 0.5 hours is taken for analysis.
- Sky size resolution – 4000 – this factor defines the scanning size of the file which needs to be scanned, if we increase the value one pixel would be analyzed more accurately.
- Mask – Building_Footprint –For processing on target buildings.
- Snap Raster – DSM
- Diffuse Proportion- This represents the proportion of global normal radiation flux that is. diffused. A value of 0.3 represents clear sky conditions. The diffuse proportion formula is as follows:
  $$Diffuse\ Proportion = Diffused\ radiation\ value\ /Global\ radiation\ value$$

- Transmissivity: represents a fraction of radiation that passes through the atmosphere. A default value of 0.5 is taken for a generally clear sky. Transmissivity is inversely proportional to diffuse proportion value.

Based on IMD global radiation and diffusion radiation data, month-wise **diffuse proportion and transmissivity values** are calculated as per Table 3 over the study regions. Calculated diffused and transmissivity values were further used for the calculation of parameters for electricity production during different weather conditions.

| LOCATION | MONTH | DIFFUSED VALUE | TRANSMITIVITY VALUE |
|---|---|---|---|
| AHMEDABAD | Jan-21 | 0.494 | 0.362 |
| | Feb-21 | 0.329 | 0.479 |
| | Mar-21 | 0.355 | 0.460 |
| | Apr-21 | 0.409 | 0.422 |
| | May-21 | 0.500 | 0.358 |
| | Jun-21 | 0.604 | 0.283 |
| GANDHINAGAR | Jul-21 | 0.682 | 0.228 |
| | Aug-21 | 0.680 | 0.230 |
| | Sep-21 | 0.600 | 0.287 |
| DELHI | Jan-21 | 0.785 | 0.155 |
| | Feb-21 | 0.850 | 0.109 |
| | Mar-21 | 0.704 | 0.212 |

*Table -3: Calculated diffusion values and transmissivity values*

### 4.4. Criteria for filtering of suitable rooftops

Acquiring solar radiation value is not just enough, we need to analyze which areas are suitable for applying solar panels, for the same we need to do parametric filtering to filter out suitable cell values. The important parameters considered for the suitable rooftops are as follows:

• Suitable rooftops should have at least $45^0$ slopes, as a greater slope may decrease the solar radiation intensity.
• Suitable rooftops should at least receive 800KWh/m$^2$ annually or 50KWh/m$^2$ monthly which will have an ideal state tendency to apply solar panels.
• Suitable Rooftops should not be facing north as north-facing rooftops receive less Sunlight. (Nanda et al, 2020)

### 4.5. Calculation of Power per Building
### 4.5.1. Aggregate cells by building

The zone is an area defined for consideration of study such as building footprint shapefile. Zonal statistics is used to generate the value of a building/ zone.

The input parameter is defined as follows:
- Input file – Building Footprint
- Zone File – Building_ID / osm_ID
- Input value raster – Suitable_Cells
- Output Table File – Solar_Rad_Table

Output Table features as follows:
- Building_ID defines a unique ID to differentiate each building from the other.
- COUNT – buildings cell counts suitable for analysis.
- AREA – the area of a single building shape ($m^2$)
- MEAN – average value of solar radiation received by a single unit.

Output table contains the building id, area and mean average value of solar radiation available on the building.

### 4.5.2. Filtering of suitable buildings
Using select-by-layer attributes, further buildings having an area greater than 1500 $ft^2$ are filtered. Buildings with an area less than $1500^2$ are not considered for applying solar panels.

### 4.5.3. Calculation of usable electricity per building
To represent production of electricity, a new field as Usable_Ele is introduced in the Output table. A generalized formula is defined to estimate the electricity generated in the output of a photovoltaic system which is as follows:

$$E = A * H * r * PR \qquad \ldots Eq\ 4$$

Where,
A – area of building rooftop ($m^2$)
r – solar panel yield percentage (%)
H – values of annual/monthly solar radiation (Kwh/$m^2$)
PR – performance ratio.

According to the Ministry of New and Renewable Energy Government of India Polycrystalline solar panels are the most widely used in India. They are made up of silicon crystal fragments. They have an efficiency of 13% to 15%. According to the PVsyst simulation, the floating PV system has a performance ratio of 76.39% and the rooftop PV system has a performance ratio of 82.69%. The floating PV system has a performance ratio of 80.24%, while the rooftop PV system has a performance ratio of 73.41% based on mathematical calculations (MNRE website).

So, the refined and finally implemented calculation formula for Electricity Production is:

$$E = A * H * 0.15 * 0.8 \qquad \ldots Eq\ 5$$

Based on the above filtering criteria's and calculations, power per building is calculated for all the suitable buildings.

## 5 Results
It was observed that the slope calculation of the 2D Raster was giving flat slope results as the 2D raster doesn't feature the height values, whereas on the other hand 3D raster features height value in "Z" coordinate which facilitates slope calculation to measure slope and aspect more accurately. So, for the same reason, the parameters (refer to Table 2) are tuned in a way to obtain suitable results.

| 2D Raster parameters | | |
|---|---|---|
| **Parameter** | **Factor to analyze** | **Value condition** |
| Remove area with higher slope | Slope Raster | Value <= 90 |
| Remove areas with lower solar radiation | Solar Radiation Raster | Value >= 45 |
| Remove areas that face north | Slope Raster | Value <= 90 |
| | Aspect Raster | 22.5 >=Value <= 337.5 |

*Table 4. 2D Raster parameters*

| 3D Raster parameters | | |
|---|---|---|

| Parameter | Factor to analyze | Value condition |
|---|---|---|
| Remove area with higher slope | Slope Raster | Value <= 45 |
| Remove areas with lower solar radiation | Solar Radiation Raster | Value >= 45 |
| Remove areas that face north | Slope Raster | Value <= 10 |
|  | Aspect Raster | 22.5 >=Value <= 337.5 |

*Table 5. 3D Raster parameters*

Results of all datasets i.e., 1 meter (2D image) & 1m, 10m and 30m (3D) over regions of Ahmedabad, Gandhinagar, and Delhi are computed and shown as per Fig 3, 4, 5 respectively. Thereafter, a comparative analysis is performed from January to June, July to September, and October to December over Ahmedabad, Gandhinagar, and Delhi respectively (Table 6,7 and 8). Fig. 6 contains the plot representation of solar power over Ahmedabad using various resolution of data. Table 9 contains the no of building best suited for solar energy whereas Table 10 lists the top no of buildings suitable for solar energy for the studied cities.

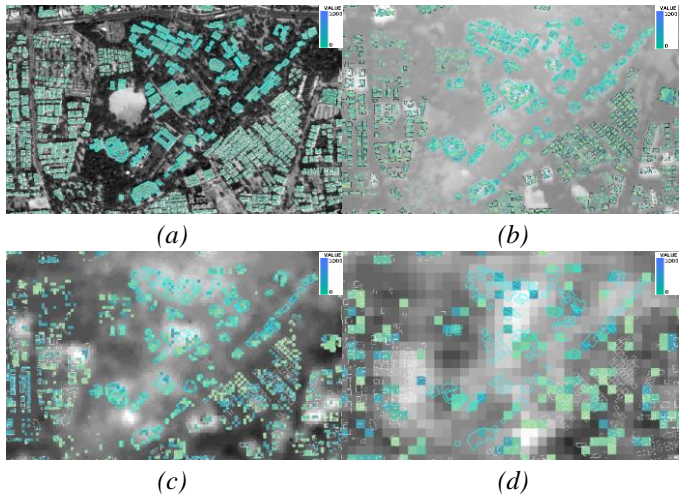

*(a)* *(b)*
*(c)* *(d)*

*Fig. 3. Ahmedabad Patch-1 results over 2D and varied res.l 3D data*

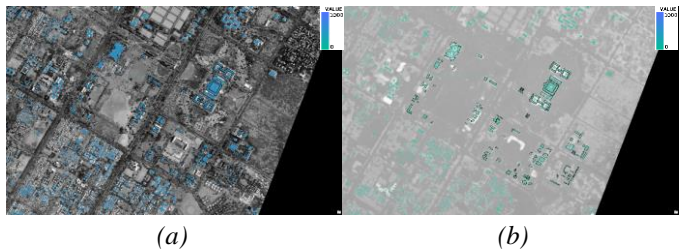

*(a)* *(b)*

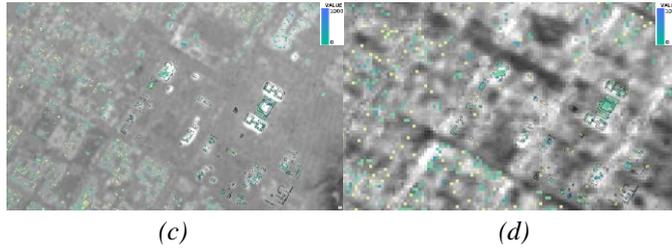

*(c)* *(d)*

Fig. 4. Gandhinagar Patch-1 results over 2D and varied res.l 3D data

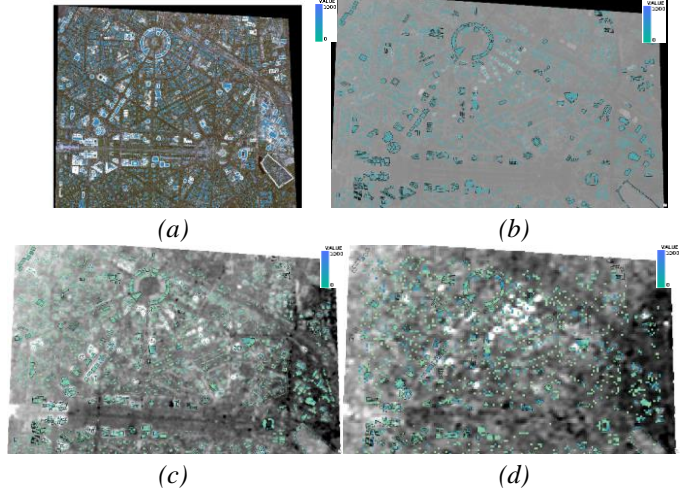

*(a)* *(b)*

*(c)* *(d)*

Fig. 5. Delhi Patch-1 results over 2D and varied res.l 3D data

Fig. 3,4 & 5, the results are captured based on different resolution of datasets viz. 2D, 1M DEM, 10m DEM and 30m DEM respectively over Ahmedabad, Gandhinagar & Delhi respectively, where Fig. (a), (b), (c), and (d) are delineated as follows:
(a) 1-meter 2-dimensional raster, which regrettably lacks specific building height data, therefore limiting its efficacy in capturing the intricate nuances of urban landscapes.
(b) 1-meter 3-dimensional Digital Elevation Model (DEM) raster, enriched with detailed building height information, thereby affording a more comprehensive depiction of urban terrain.
(c) 10-meter 3D DEM raster, albeit encompassing building height data, operates at a coarser resolution, with a loss of granularity in shade analysis and rooftop orientation.
(d) A 30-meter 3D DEM raster, possessing building height data at a notably reduced resolution, further exacerbates the amalgamation of multiple structures into indistinct entities, thereby complicating the accurate assessment of solar energy potential.

The analysis underscores a critical distinction among the visualizations presented. Fig 3(a), 4(a), 5(a) depict solar radiation rasters sans height values, operating under the assumption of uniform building heights. This approach, while practical, introduces inaccuracies in shade analysis and the determination of rooftop orientation. Conversely, Fig 3(c), 4(c), 5(c) and Fig 3(d), 4(d) and 5(d) portray 3-dimensional solar radiation rasters with resolutions of 10m and 30m respectively.

However, these representations amalgamate multiple buildings into singular entities, thereby obscuring the precise delineation of individual structures and complicating the assessment of electricity production potential.

In contrast, Fig 3(b), 4(b), 5(b) emerge as the most adept visualization for meticulous analysis. Boasting a superior resolution of 1m and comprehensive height data, this representation offers a detailed of the urban landscape. By incorporating precise height information, Fig 3(b), 4(b), 5(b) facilitate a more accurate assessment of solar shading patterns and rooftop orientations, thereby enhancing the reliability of predictions regarding electricity production potential. Consequently, it stands out as the optimal choice for comprehensive and meticulous analysis in this context.

Table 6 denotes the electricity production values calculated over Ahmedabad from January to June in ideal conditions and calibrated conditions. Ideal conditions are defined, where the values of solar diffusion and transmissivity are not considered whereas, calibrated conditions consider the IMD data of cloud cover, rain, diffusions, and transmissivity. In this table "OSM ID" refers to unique identification numbers given to each building and "H" denotes the elevation of the building in meters. We can observe that the radiation values are gradually increasing from January to May and drop in the month of June, this happens due to the presence of rainfall affecting the incoming Sunrays due to cloudy weather.

Whereas, Tables 7 & Table 8 denote the electricity production values calculated over Gandhinagar from June to September and Delhi from January to March respectively January to June in an ideal condition and calibrated condition in this table "OSM ID" refers to unique identification numbers given to each building and "H" denotes the elevation of the building in meters. Table 7, Over Gandhinagar, it is observed that the radiation values are gradually decreasing from July to September, which occurs due to the presence of rainfall affecting the incoming Sunrays due to cloudy weather. Over Delhi, due to foggy weather in January, lower values are observed in Jan and Feb whereas an increase in values of electricity production is noticed in March.

| Dataset Features | | | Ideal Condition Electricity Production value (Kwh) in 2021 | | | | | | Calibrated Condition Electricity Production value (Kwh) in 2021 | | | | | |
|---|---|---|---|---|---|---|---|---|---|---|---|---|---|---|
| OSM ID | H | Area (sqm) | Jan | Feb | Mar | Apr | May | Jun | Jan | Feb | Mar | Apr | May | Jun |
| **367546649** | 19 | 3828.709 | 36268 | 40901 | 57051 | 65167 | 72863 | 71951 | 28254 | 39657 | 54171 | 59348 | 59348 | 54705 |
| **367546599** | 15 | 1208.663 | 11085 | 12553 | 17971 | 20827 | 23466 | 23255 | 8776 | 12189 | 17097 | 18980 | 18980 | 17653 |
| **367546613** | 14 | 965.1673 | 8604 | 9604 | 13646 | 15948 | 18085 | 17977 | 6943 | 9330 | 12995 | 14545 | 14545 | 13632 |
| **367552933** | 14 | 1596.283 | 15179 | 17174 | 24079 | 27239 | 30200 | 29716 | 11787 | 16645 | 22868 | 24811 | 24811 | 22687 |
| **367553070** | 13 | 715.7119 | 7460 | 8001 | 10422 | 10925 | 11388 | 10882 | 5532 | 7708 | 9832 | 9910 | 9910 | 8457 |
| **5463891** | 12 | 7125.598 | 70813 | 79166 | 109530 | 122967 | 135470 | 132903 | 54450 | 76657 | 103934 | 111939 | 111939 | 101718 |
| **366622338** | 11 | 940.928 | 8187 | 9338 | 13362 | 15475 | 17487 | 17354 | 6661 | 9097 | 12725 | 14129 | 14129 | 13232 |
| **5463889** | 11 | 1266.182 | 11343 | 12191 | 17141 | 20014 | 22654 | 22489 | 9058 | 11852 | 16301 | 18213 | 18213 | 16970 |
| **367546609** | 10 | 854.6821 | 7120 | 7884 | 11587 | 13888 | 16027 | 16049 | 5863 | 7704 | 11070 | 12703 | 12703 | 12173 |
| **367546598** | 9 | 681.1754 | 6613 | 7465 | 10464 | 11821 | 13080 | 12857 | 5169 | 7233 | 9934 | 10764 | 10764 | 9828 |
| **5463890** | 8 | 1398.511 | 13089 | 14911 | 21173 | 24212 | 27025 | 26666 | 10215 | 14455 | 20100 | 22013 | 22013 | 20193 |
| **367546741** | 7 | 362.9097 | 3501 | 3898 | 5254 | 5976 | 6634 | 6531 | 2756 | 3776 | 4992 | 5447 | 5447 | 5003 |
| **367546607** | 5 | 336.0162 | 2897 | 3404 | 4979 | 5855 | 6670 | 6642 | 2343 | 3309 | 4743 | 5337 | 5337 | 5017 |

*Table 6. Electricity Production of Patch 1 Ahmedabad Month-wise Ideal & Calibrated Condition*

| Dataset Features | | | Ideal Condition Electricity Production value (Kwh) in 2021 | | | Calibrated Condition Electricity Production value (Kwh) in 2021 | | |
|---|---|---|---|---|---|---|---|---|
| OSM ID | Floors | Area (sqm) | Jul-21 | Aug-21 | Sep-21 | Jul-21 | Aug-21 | Sep-21 |
| 4E169A71C6A46D43-135-0 | 20 | 1530.05 | 31453 | 29687 | 24933 | 21935 | 20595 | 18510 |
| 2D190A71C6A46D43-81-0 | 19 | 1208.65 | 24846 | 23451 | 19695 | 17327 | 16269 | 14622 |
| 84B6D271C6A46D43-292-0 | 18 | 1400.1 | 28782 | 27166 | 22815 | 20072 | 18846 | 16938 |
| 49B4E171C6A46D43-114-0 | 17 | 844.43 | 17359 | 16384 | 13760 | 12106 | 11366 | 10215 |
| 49B4E171C6A46D43-114-0 | 17 | 844.43 | 17359 | 16384 | 13760 | 12106 | 11366 | 10215 |
| 6DB29A71C6A46D43-349-0 | 16 | 1399.57 | 28771 | 27156 | 22806 | 20064 | 18839 | 16931 |
| 7142E171C6A46D43-307-0 | 15 | 407.86 | 8384 | 7914 | 6646 | 5847 | 5490 | 4934 |
| 2914E171C6A46D43-80-0 | 14 | 493.48 | 10145 | 9575 | 8041 | 7075 | 6642 | 5970 |
| 89A72171C6A46D43-330-0 | 14 | 2446.17 | 50286 | 47462 | 39861 | 35068 | 32926 | 29593 |
| 70C29371C6A46D43-282-0 | 11 | 652.44 | 13412 | 12659 | 10632 | 9353 | 8782 | 7893 |
| 48B8E271C6A46D43-90-0 | 11 | 1120.99 | 23044 | 21750 | 18267 | 16070 | 15089 | 13561 |
| 8898CC71C6A46D43-238-0 | 10 | 1516.39 | 31172 | 29422 | 24710 | 21739 | 20411 | 18345 |
| 48C2D471C6A46D43-12-0 | 9 | 249.23 | 5123 | 4836 | 4061 | 3573 | 3355 | 3015 |
| 2CC2CC71C6A46D43-124-0 | 9 | 930.46 | 19128 | 18053 | 15162 | 13339 | 12524 | 11256 |
| 2D145471C6A46D43-61-0 | 8 | 194.96 | 4008 | 3783 | 3177 | 2795 | 2624 | 2359 |
| 8DA2CC71C6A46D43-213-0 | 6 | 244.07 | 5017 | 4736 | 3977 | 3499 | 3285 | 2953 |

*Table 7. Electricity Production (Kwh) of Patch 1 Gandhinagar Month-wise Ideal & Calibrated Condition*

| Dataset Features | | | Ideal Condition Electricity Production value (Kwh) Values | | | Calibrated Electricity Production value (Kwh) Values | | |
|---|---|---|---|---|---|---|---|---|
| OSM ID | H | Area (sqm) | Jan-21 | Feb-21 | Mar-21 | Jan-21 | Feb-21 | Mar-21 |
| 53549661 | 27 | 24332.8 | 185809 | 229020 | 350054 | 71257 | 86757 | 215861 |
| 53549662 | 24 | 24303.2 | 185583 | 228742 | 349629 | 71170 | 86651 | 215599 |
| 53549663 | 22 | 18114.29 | 138323 | 170492 | 260594 | 53046 | 64585 | 160696 |
| 53549664 | 20 | 11443.7 | 87386 | 107708 | 164630 | 33512 | 40802 | 101519 |
| 53549665 | 19 | 14233.58 | 108690 | 133967 | 204766 | 41682 | 50749 | 126269 |
| 53549666 | 18 | 8047.2 | 61450 | 75740 | 115768 | 23566 | 28692 | 71388 |
| 53549667 | 17 | 6575.65 | 50213 | 61890 | 94598 | 19256 | 23445 | 58334 |
| 53549668 | 15 | 8791.48 | 67133 | 82745 | 126475 | 25745 | 31345 | 77991 |
| 53549669 | 13 | 5081.4 | 38802 | 47826 | 73102 | 14880 | 18117 | 45078 |
| 53549670 | 11 | 1226.24 | 9364 | 11541 | 17641 | 3591 | 4372 | 10878 |
| 53549671 | 10 | 818.15 | 6248 | 7700 | 11770 | 2396 | 2917 | 7258 |
| 53549672 | 9 | 5947.16 | 45413 | 55975 | 85557 | 17416 | 21204 | 52759 |
| 53549673 | 9 | 3314.82 | 25312 | 31199 | 47687 | 9707 | 11819 | 29406 |
| 53549674 | 8 | 1016.81 | 7765 | 9570 | 14628 | 2978 | 3625 | 9020 |
| 53549675 | 7 | 1506.97 | 11507 | 14184 | 21679 | 4413 | 5373 | 13369 |
| 53549676 | 6 | 157.01 | 1199 | 1478 | 2259 | 460 | 560 | 1393 |
| 53549677 | 5 | 8393.84 | 64097 | 79003 | 120755 | 24581 | 29928 | 74463 |
| 53549678 | 5 | 639.58 | 4884 | 6020 | 9201 | 1873 | 2280 | 5674 |
| 53549679 | 2 | 168.7 | 1288 | 1588 | 2427 | 494 | 602 | 1497 |
| 53549680 | 2 | 172.35 | 1316 | 1622 | 2479 | 505 | 614 | 1529 |
| 53549681 | 1 | 208.48 | 1592 | 1962 | 2999 | 611 | 743 | 1849 |

*Table 8. Electricity Production (Kwh) of Patch 1 Delhi Month-wise Ideal & Calibrated Condition*

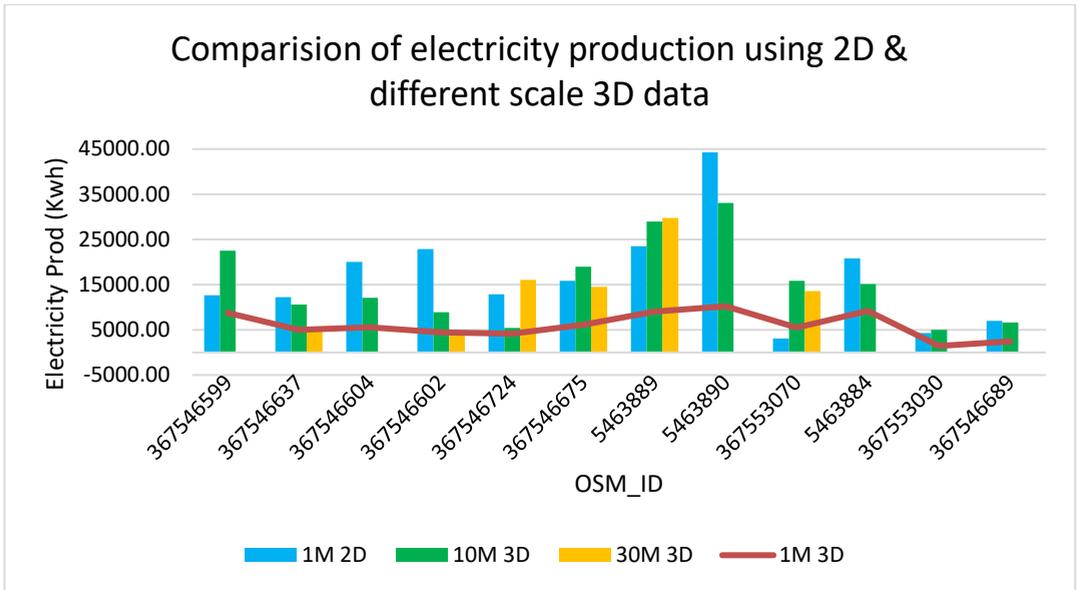

*Fig. 6. Comparison of electricity production over Ahmedabad using 2D & different scale 3D data.*

Refer Fig 6, further analysis is performed over Ahmedabad by comparing results derived from various datasets, including both 2D and 3D data at different resolutions. When examining the 2D data at a 1-meter resolution, it became evident that the analysis solely focused on the rooftop area for calculating electricity production, assuming all buildings were at uniform heights. In contrast, the 3D datasets, with resolutions of 1m, 10m, and 30m, considered the heights of buildings and terrain. This inclusion enabled a more comprehensive analysis, particularly regarding shade and rooftop orientation. However, we encountered a significant issue with the 10m and 30m resolution data, where clusters of buildings (2-3 buildings for 10m and 4-8 buildings for 30m) were amalgamated and treated as single rooftops. Consequently, unique values were calculated for these merged patches, potentially skewing the results. Our observations suggest that both 2D data and 30m resolution 3D data are better suited for assessing solar potential across extensive areas, such as entire states or multiple states. Meanwhile, the 10m resolution 3D data proves more valuable for analysing large IT industry zones characterized by expansive rooftops. Notably, the 1m resolution 3D data stands out for its ability to capture variations in building heights and Sun positions, thus offering greater authenticity in the analysis. This level of detail enhances the precision of solar potential assessments, particularly in urban environments where building heights vary significantly.

| Top locations to apply solar panels | |
|---|---|
| **City Name** | **No of buildings** |
| **Ahmedabad** | 363 |
| **Gandhinagar** | 609 |
| **Delhi** | 780 |

*Table 9. Top locations to apply solar panels in Ahmedabad, Gandhinagar, and Delhi*

In the context of urban development, it is imperative to identify buildings that are conducive to harnessing solar energy. This not only aids in cost reduction associated with architecture but also optimizes land usage within urban areas. Table 9 provides a breakdown of the number of buildings within Ahmedabad, Gandhinagar, and Delhi that are deemed suitable for the installation of solar panels.

| TOP 10 BUILDINGS COORDINATES TO APPLY SOLAR PANEL | | | |
|---|---|---|---|
| City Name | Building OSM ID/UID | Y Coordinate | X Coordinate |
| Ahmedabad | 366439316 | 72.5240012 | 23.0386721 |
| | 366439322 | 72.5236082 | 23.0377847 |
| | 366439323 | 72.5241848 | 23.0379489 |
| | 367949980 | 72.4914164 | 23.0207438 |
| | 367949985 | 72.4950372 | 23.0197335 |
| | 367949990 | 72.4930263 | 23.0200307 |
| | 367950021 | 72.4909116 | 23.0212890 |
| | 366439711 | 72.4944626 | 23.0405740 |
| | 367950032 | 72.4899761 | 23.0195591 |
| | 366452914 | 72.5265379 | 23.0359598 |
| Gandhinagar | 31C88971C6A46D43-122-0 | 72.6466014 | 23.2224688 |
| | 31C88971C6A46D43-122-0 | 72.6466014 | 23.2224688 |
| | 85949A71C6A46D43-237-1 | 72.6568930 | 23.2191112 |
| | 85949A71C6A46D43-237-1 | 72.6568930 | 23.2191112 |
| | 85949A71C6A46D43-375-0 | 72.6568451 | 23.2191005 |
| | 85949A71C6A46D43-375-0 | 72.6568451 | 23.2191005 |
| | 2D36E431C6A46D43-185-0 | 72.6584687 | 23.2255352 |
| | 2D36E431C6A46D43-185-0 | 72.6584687 | 23.2255352 |
| | 72169A71C6A46D43-225-0 | 72.6558999 | 23.2176203 |
| | 4D290A71C6A46D43-175-0 | 72.6578694 | 23.2207361 |
| Delhi | 53549661 | 77.2057291 | 28.6129585 |
| | 53549662 | 77.2058994 | 28.6151688 |
| | 53549663 | 77.2342706 | 28.6245876 |
| | 53549664 | 77.2252423 | 28.5895973 |
| | 53549665 | 77.2081478 | 28.6171623 |
| | 53549666 | 77.2073200 | 28.6187148 |
| | 53549667 | 77.2434507 | 28.6158335 |
| | 53549668 | 77.2033659 | 28.6160416 |
| | 53549669 | 77.2151134 | 28.6110651 |
| | 53549670 | 77.2170234 | 28.6328189 |

*Table 10. Coordinates of top 10 Ideal locations to apply solar panels in Ahmedabad, Gandhinagar, and Delhi*

Meanwhile, Table 10 highlights the top 10 buildings across these cities that exhibit significant potential for solar panel application, based on criteria such as available area and capacity to capture solar radiation. These tables serve as valuable tools for city planners and developers, offering insights into which buildings hold promise for integrating solar energy infrastructure. By leveraging such data, urban projects can prioritize sustainable practices, enhance energy efficiency, and contribute to a greener, more resilient urban landscape.

## 6 Discussions

The objective of this study is to determine the solar potential of building rooftops over the cities of Ahmedabad, Gandhinagar, and Delhi using high-resolution satellite imagery and a Digital Elevation Model. The analysis is performed on DEM of resolution 1m, 10m, and 30m along with a 2D raster of resolution 1m. The methodology consisted of various GIS techniques using ArcGIS software. It is observed that the solar radiation calculation depends on the elevation of the raster. Further, the filtering process is performed based on the slope and aspect of the raster. We have chosen Ahmedabad, Gandhinagar, and Delhi for solar potential calculation study as all these cities face high heat in summer and have variable temperatures ranging from $15^0$C to $45^0$C which makes them suitable for solar potential in different scenarios. Also, in these cities' buildings with diverse building heights and rooftop areas are available for analysis.

Solar radiation calculators such as European solar calculator and PVSyst consider 2D data only for calculation of solar radiation, therefore, resulting in a rough calculation of the values considering only the area of the building and Sun positioning (ESRI website, PVSYST website). Referring our study, In the case analysis using 3D data, the height of the building is considered for analysis which brings in a parameter such as shade analysis and direct analysis of Sun.

Fig. 3, 4, and 5 represent the differentiation of datasets based on data resolution. In these Figs, 1m 2D & 1m 3D datasets are accurately covering each building whereas 10m and 30m data are covering many buildings and depicting single value for those. By performing analysis using 30m / 10m and 1m DEM, it is observed that 1m 3D DEM compared to 10m and 30m data is much more accurate for the calculation of solar potential over an urban city. However, if solar potential on a larger area such as over a city, town, or state is required to be estimated, then 10m and 30m 3D DEM are better than only 2D image data.

Earlier, default parameters of diffusivity 0.3 and transmissivity 0.5 were used for a clear sky, which calculates solar radiation values according to ideal conditions. Thereafter, month-wise IMD ground stations' global radiation and diffusion radiation data were taken for analysis. These values were used for the calculation of the diffusion ratio and transmissivity value. As we compared the ideal values with calibrated values, we observed a percentage decrease of 5-20% in Ahmedabad from January to June 25-30% in Gandhinagar from July to September and 30-65% in Delhi from Jan to March in solar radiation values in realistic conditions. A significant fall in the percentage values i.e. 60% is observed due to fog and the winter season in the case of Delhi during January and February.

Solar energy can also be affected by various technical reasons such as accumulation of dust or dirt, angle, and orientation of solar panel, hot air release of neighboring devices such as AC duct or chimney, atmosphere temperature, reflection, glare, inverter efficiency, electrical resistance, system aging and degradation, and grid-related loss. Due to environmental and technical reasons as listed above, there can be a loss of up to 50-55% of electricity production using solar panels.

## 7  Conclusion and Future Scope

In this study, solar potential analysis using high-resolution satellite imagery (50cm) and DEM (1m/10m/30m) over urban cities was explored. Using high-resolution data and derived diffusion and transmissivity values from IMD data, the solar potential of building rooftops was calculated over the city of Ahmedabad, Gandhinagar, and Delhi for different seasons. A parametrical and comparative analysis is performed where parameters such as atmospheric conditions, rooftop slope, direction, shade analysis, and area rectification are done to filter the accurate results. For estimating the solar potential of a rooftop, there are various online services and software, but the solar potential is calculated using only 2-dimensional data. Solar radiation values using 2D data can only provide solar potential based on area and Sun position. 10m / 30m DEM-based 3D data can be useful in estimating the solar potential over larger areas. Whereas, using 3D 1m data, more accurate and differentiable results can be achieved. It is concluded that solar radiation values during realistic weather conditions as compared to ideal/clear sky conditions, may result in 20-50% variations. It is further observed that due to environmental and technical factors, there can be a cumulative loss of up to 50-55% of electricity production using solar panels. As a part of future scope, we will be validating our results, using the ground data.